\documentclass[twoside]{article}
\usepackage[accepted]{aistats2022}
\usepackage[round]{natbib}

\usepackage{amsthm,xcolor,bbm,amsfonts}
\usepackage[colorlinks=true,citecolor=blue]{hyperref}
\usepackage{cleveref}
\usepackage{subfig}
\usepackage{graphicx}

\theoremstyle{plain}  
\newtheorem{thm}{Theorem}[section]

\crefname{assumption}{Assumption}{Assumptions}
\crefname{equation}{Eq.}{Eqs.}
\crefname{figure}{Fig.}{Figs.}
\crefname{table}{Table}{Tables}
\crefname{section}{Sec.}{Secs.}
\crefname{theorem}{Thm.}{Thms.}
\crefname{lemma}{Lemma}{Lemmas}
\crefname{corollary}{Cor.}{Cors.}
\crefname{example}{Example}{Examples}
\crefname{appendix}{Appendix}{Appendixes}
\crefname{remark}{Remark}{Remark}

\begin{document}

\runningauthor{Badr-Eddine Chérief-Abdellatif, Yuyang Shi, Arnaud Doucet, Benjamin Guedj}

\twocolumn[

\aistatstitle{On PAC-Bayesian reconstruction guarantees for VAEs}

\aistatsauthor{Badr-Eddine Chérief-Abdellatif \And Yuyang Shi}
\aistatsaddress{University of Oxford \And University of Oxford}

\aistatsauthor{Arnaud Doucet \And  Benjamin Guedj}
\aistatsaddress{University of Oxford  \And University College London and Inria}
]

\begin{abstract}
Despite its wide use and empirical successes, the theoretical understanding and study of the behaviour and performance of the variational autoencoder (VAE) have only emerged in the past few years. We contribute to this recent line of work by analysing the VAE's reconstruction ability for unseen test data, leveraging arguments from the PAC-Bayes theory. We provide generalisation bounds on the theoretical reconstruction error, and provide insights on the regularisation effect of VAE objectives. We illustrate our theoretical results with supporting experiments on classical benchmark datasets.
\end{abstract}

\section{INTRODUCTION}

Since its introduction by \cite{kingma2013auto}, the Variational AutoEncoder (VAE) has attracted considerable interest and is now widely used for learning low dimensional representations of high dimensional complex data, such as images. The VAE provides a probabilistic view on the autoencoder, a structure which trains an encoder that maps a high dimensional input to a low dimensional latent code, which is then reconstructed using a decoder. The probabilistic encoder $q_\phi(\mathbf{z}|\mathbf{x})$ is a distribution over the possible values of the code $\mathbf{z}$ given a datapoint $\mathbf{x}$, while the probabilistic decoder $p_\theta(\mathbf{x}|\mathbf{z})$ is a distribution over the possible corresponding values of $\mathbf{x}$ given the code $\mathbf{z}$. 

As any autoencoder, the VAE offers a powerful framework for learning compressed representations by encoding the information required to reconstruct the original signal accurately. Beyond this simple coding theory perspective, the VAE is more generally presented as a deep generative model. Assuming that the latent code $\mathbf{z}$ is distributed according to a prior $p(\mathbf{z})$ (typically an isotropic Gaussian distribution) and that the decoder is defined via a likelihood $p_\theta(\mathbf{x}|\mathbf{z})$ parameterised by a neural network, the encoder $q_\phi(\mathbf{z}|\mathbf{x})$ is then represented as a variational approximation of the posterior $p_\theta(\mathbf{z}|\mathbf{x})$ which is also parameterised by a neural network. Both the encoder and the decoder networks weights are jointly learnt by minimising a variational objective:
$$
\underbrace{- \mathbb{E}_{q_{\phi}(\mathbf{z}|\mathbf{x})} \left[\log p_{\theta}(\mathbf{x}|\mathbf{z}) \right]}_{\text{reconstruction loss}} + \underbrace{\mathrm{KL}(q_{\phi}(\mathbf{z}|\mathbf{x})\Vert p(\mathbf{z}))}_{\text{rate}} 
$$
averaged over the dataset, where the first term is a reconstruction loss and the second term is the Kullback--Leibler (KL) divergence between the encoder and the prior $p(\mathbf{z})$. This can alternatively be seen as maximising the celebrated Evidence Lower Bound (ELBO) on the (intractable) log-evidence of this latent variable model. In information-theoretic words, the reconstruction loss is the distortion measured through the encoder-decoder channel, while the KL term is called the rate, an upper bound on the mutual information between the input and the code (a quantity which is usually interpreted as a regulariser controlling the degree of compression through the autoencoder). The celebrated $\beta$-VAE variant \citep{Higgins2017} corresponds to the case where the KL rate has a multiplicative factor $\beta$ in the variational objective.

\textbf{Related work.}
There is a growing body of works aiming to understand the empirical success of the VAE, and a number of reasons have been put forward to explain its apparent good generalisation properties. A large part of the literature has addressed generalisation through the lens of generative modeling, by measuring the sampling ability of the VAE and of its variants using either the log-marginal likelihood, the ELBO, or relative quantitative metrics such as FID \citep{HeuselFID2017,KumarPoole2020} and precision and recall scores \citep{Sajjadi2018}. Rate-distortion curves rather than the log-likelihood have also been proposed to obtain more information about the model \citep{Alemi2018,Huang2020}. Probably the most remarkable phenomenon is the now widely acknowledged fact that an infinite capacity VAE memorises the training data and interpolates: in other words the VAE predicts novel data points between given training samples by decoding a convex combination of the latent codes \citep{Alemi2018,RezendeViola2018,Shu2018}. In light of these findings, \cite{Shu2018} investigated the impact of the encoder capacity on the memorisation property, while \cite{Zhao2018} focused on the question of sampling out-of-domain data from the learnt representation. Another interesting line of research lies in evaluating the quality of the representation via different semantic notions such as disentanglement (achieving interpretability via the decomposability of the latent representation, with the hope to ultimately generalise to new combinations of factors, as explored by \citealp{Higgins2017,Chen2018,KimMnih2018,Emile2019,Esmaeili2019,Locatello2019}) or robustness (by exploring metrics that capture some effects that rare events from multiple generative factors can have on feature encodings, see \citealp{Suter2019}). While this is out of scope of the present paper, we also acknowledge recent works on other generative models, such as Generative Adversarial Networks \citep{DBLP:journals/jmlr/BiauST21,DBLP:conf/alt/SchreuderBD21}.

Nevertheless, the notion of generalisation is intrinsically subjective. Indeed, a given VAE objective can lead to good reconstruction on unseen test data while being poor for sampling, and can conversely lead to poor reconstruction while being able to generate realistic images. Furthermore, while the ELBO objective naturally defines a proper generative model as a lower bound on the log-marginal likelihood, this is no longer true for the $\beta$-VAE when $\beta<1$. Hence, the approach consisting in evaluating any VAE objective from the sampling perspective is not always appropriate. We do not focus here on the generative abilities of the VAE. Similarly, although disentanglement potentially induces generalisation, we are not focusing on that notion per se.

In this paper, we study the VAE from a reconstruction perspective: we consider the VAE as a model that learns a lossy encoder and decoder, with the belief that a model generalising well should capture a meaningful representation of the data.
To better understand the generalisation ability of the VAE in terms of reconstruction, \cite{Bozkurt2021} have investigated the regularisation properties of the VAE objective. Somewhat counter-intuitively, they demonstrated through extensive experiments that the KL term neither actually acts as a regulariser nor improves generalisation when focusing on reconstruction. They also showed that reducing $\beta$ always decreases the generalisation gap when test data deviates substantially from the training data in pixel space. Their work differentiates between test data that can be reconstructed easily by taking the most similar memorised training data points, and more complicated test data with out-of-domain samples. Hence, the influence of the KL term on the generalisation ability of the VAE is not the same depending on the difficulty of the generalisation task, although the impact of the KL is always monotonic in $\beta$. This sheds additional light on the observations made by \cite{Alemi2018} and \cite{RezendeViola2018} which were conducted on training data only. This somewhat surprising behaviour calls for a further study of the regularisation effect of VAE objectives -- a contribution of the present paper.

\textbf{Our approach.}
A natural way to study regularisation is to derive statistical guarantees to quantify the risk of overfitting. We address this by computing generalisation bounds on the reconstruction loss using PAC-Bayes theory (pioneered by \citealp{STWilliamson1997,McAllester1999,Seeger2002,McAllester2003,Maurer2004,CatoniThermo} among others -- we refer to \citealp{guedj2019primer} and \citealp{alquier2021user} for recent surveys). PAC-Bayes has been extensively and successfully used in many settings in machine learning and statistics -- however, to the best of our knowledge, it has never been leveraged in the VAE literature. The inference model represented by the encoder is a stochastic function of the inputs that is learnt using amortised inference to improve computational efficiency for huge datasets.

We first formulate PAC-Bayes bounds for the VAE structure. We then show that minimising directly the PAC-Bayes bound over the reconstruction error for amortised variational inference not only provides non-vacuous generalisation bounds, but also significantly decreases the generalisation gap between the test and the training reconstruction errors.
The idea of using PAC-Bayes to evaluate the generalisation ability of autoencoders has appeared in the past few years. \cite{EpsteinMeir2019} have indeed recently adapted margin- and norm-based results for deep neural networks \citep{Bartlett2017,Neyshabur2018,Arora2018} to obtain a generalisation bound for deterministic autoencoders. However, there are two substantial differences between their work and ours. First, their bound on the generalisation gap can only be obtained up to a large constant independent of the network parameters, the sample size and the margin, while our bound can be computed and used as an objective for designing an alternative learning algorithm. Second, their bound is deterministic. This is due to the fact that PAC-Bayes inequalities only appear as an artefact in their proofs, in which the networks parameters are artificially perturbed using a Gaussian noise. The deterministic generalisation gap is then controlled by the means of the perturbed network parameters at the price of a looser inequality involving different margin levels. In contrast, our bound is a genuine PAC-Bayes bound on the probabilistic structure of the VAE whose stochasticity is used as a way to inject noise during the learning phase.

\textbf{Summary of our contributions.}
Hence, the primary motivation for this work is to
complement the findings on the role of the rate as a regulariser \cite{Bozkurt2021} by providing the first theoretical results on the generalisation ability of the VAE and the regularising property of the KL in terms of reconstruction. 
We choose to derive statistical guarantees by computing generalisation bounds on the reconstruction loss. Leveraging PAC-Bayes theory, we provide bounds that can not only be computed empirically, but can also be used as new learning objectives with good generalisation properties and strong theoretical guarantees. Consequently, our contribution is two-fold: i) we formulate a derandomised PAC-Bayes generalisation bound for the VAE structure which is the first such bound in the VAE literature; ii) we use a non-derandomised variant of this bound to propose a novel PAC-Bayes objective for the VAE structure that will generalise well while achieving tight risk certificates. We provide empirical evidence on real-world datasets, evaluate the generalisation ability of both the $\beta$-VAE (including the original VAE with $\beta=1$) and PAC-Bayes objectives, and compute generalisation bounds for these strategies.

\section{NOTATION AND BACKGROUND}\label{sec:notation}

We consider a dataset $\mathcal{S}=\{\mathbf{x}_1$,\dots,$\mathbf{x}_n\}$ of independent copies of a random variable $\mathbf{x}\in\mathcal{X}\subset \mathbb{R}^D$ sampled from an unknown probability distribution $\mathcal{D}$, with $D$ a (potentially large) positive integer.
We assume a generative model $p_{\theta}(\mathbf{x},\mathbf{z})$ involving a latent random variable $\mathbf{z}$ in a lower dimensional space $\mathcal{Z}\subset\mathbb{R}^d$: the model is composed of a prior $p(\mathbf{z})$ (\emph{e.g.} a standard Gaussian distribution), and of a conditional likelihood $p_{\theta}(\mathbf{x}|\mathbf{z})$ from a parametric family indexed by $\theta\in\Theta$ (\emph{e.g.} the weights of a neural network). The marginal likelihood over the observed variables, given by $p_{\theta}(\mathbf{x}) = \int p_{\theta}(\mathbf{x}|\mathbf{z}) p(\mathbf{z}) d\mathbf{z}$, is typically intractable.

\subsection{Variational AutoEncoders}

The VAE \citep{kingma2013auto} adopts a variational approach to turn the intractable posterior inference and learning problem into a tractable one, which results in the maximisation of a lower bound on the log-evidence (ELBO). The encoder and the decoder, respectively parameterised by $\phi$ and $\theta$, attempt to learn: (i) a variational distribution $q_{\phi}(\mathbf{z}|\mathbf{x})$ that approximates the intractable posterior distribution $p_{\theta}(\mathbf{z}|\mathbf{x})$, and (ii) the conditional likelihood $p_{\theta}(\mathbf{x}|\mathbf{z})$ that approximates the data generating distribution. We recall the variational objective minimised by the $\beta$-VAE \citep{Higgins2017}, an extension of the VAE that reweights the KL term in the variational objective:
\begin{align*}
\mathcal{L}_{\beta}(\mathbf{\phi},\mathbf{\theta}) = \sum_{i=1}^n - & \mathbb{E}_{q_{\phi}(\mathbf{z}_i|\mathbf{x}_i)} \left[\log p_{\theta}(\mathbf{x}_i|\mathbf{z}_i) \right] \\ & + \beta \cdot\sum_{i=1}^n \mathrm{KL}(q_{\phi}(\mathbf{z}|\mathbf{x}_i)\Vert p(\mathbf{z})) .
\end{align*}
The standard VAE framework corresponds to $\beta=1$, in which case the variational objective can be rewritten as the (opposite of the) ELBO:
\begin{align*}
\mathcal{L}_1(\mathbf{\phi},\mathbf{\theta}) & = \sum_{i=1}^n - \log p_{\theta}(\mathbf{x}_i) + \sum_{i=1}^n \mathrm{KL}(q_{\phi}(\mathbf{z}|\mathbf{x}_i)\Vert p_{\theta}(\mathbf{z}|\mathbf{x}_i)) \\ & \geq \sum_{i=1}^n -\log p_{\theta}(\mathbf{x}_i) . 
\end{align*}

The prior over the latent variables is typically set to be the isotropic multivariate Gaussian $p_{\theta}(\mathbf{z})=\mathcal{N}(\mathbf{0,I}_d)$, while the conditional likelihood $p_{\theta}(\mathbf{x}|\mathbf{z})$ is generally defined as a Gaussian (in case of real-valued data) or Bernoulli (in case of binary data) whose distribution parameters are computed from $\mathbf{z}$ using a neural network. For binary data $\mathbf{x}$ for instance, the shape of the variational and likelihood distributions can be taken as a Gaussian latent distribution and a factorised Bernoulli observation likelihood:
\begin{equation}
\label{variational}
q_{\phi}(\mathbf{z}|\mathbf{x}) = \mathcal{N}\left(\mathbf{z};\boldsymbol{\mu}_\phi(\mathbf{x}), \textrm{diag}(\boldsymbol{\sigma}^2_\phi(\mathbf{z}))\right),
\end{equation}
\begin{multline}
\label{model}
\log p_{\theta}(\mathbf{x}|\mathbf{z}) = \sum_{j=1}^D \{ x_j \log \omega_\theta(\mathbf{z})_j \\
\quad \quad \quad + (1-x_j) \log (1-\omega_\theta(\mathbf{z})_j) \} ,
\end{multline}
where both the encoder distribution parameters $(\boldsymbol{\mu}_\phi(\mathbf{x}),\log\boldsymbol{\sigma}_\phi(\mathbf{x}))=\textnormal{NN}_{\phi}(\mathbf{x})$ and the decoder distribution parameter $\boldsymbol{\omega}_\theta(\mathbf{z})=\textnormal{NN}_{\theta}(\mathbf{z})$ are outputs of neural networks, with $0<\omega_\theta(\mathbf{z})_j<1$ for any $j$, which can be obtained for example via a sigmoid nonlinearity as the last layer of the neural network. Here, $\phi$ and $\theta$ are the weights of the corresponding neural networks.

Let us stress that we focus on the generalisation properties of the VAE and its variants in terms of reconstruction, and mainly interpret the structure as a model for learning representations using an encoder and a decoder. Note that we recover the case of a deterministic autoencoder in the limit of infinite capacity when $\beta=0$ as the reconstruction loss alone is minimised when the encoder $q_{\phi}(\mathbf{z}|\mathbf{x})$ is a Dirac mass at $\arg\max_{\mathbf{z}} \log p_{\theta}(\mathbf{x}|\mathbf{z})$ and when $\theta$ minimises the corresponding log-likelihood. This partly explains the fact that the KL term is often interpreted as a regulariser that smoothes the representation and makes the VAE less prone to overfitting.

\subsection{Elements from the PAC-Bayes theory}

Consider a parametric family of hypotheses indexed by $\omega\in\Omega$ (\emph{e.g.} neural networks parameterised by their weights $\omega$) and a bounded\footnote{Note that recent works have proposed strategies to relax the boundedness assumption \citep{PierreBen2018,Holland2019,KuzborskijSzepesvari2019,GrunwaldMehta2020,Haddouche2021}.} loss function $\ell:\Omega\times\mathcal{X}\rightarrow[0,1]$. We have a training sample $\mathcal{S}=\{\mathbf{x}_1$,\dots,$\mathbf{x}_n\}$ drawn from an unknown distribution $\mathcal{D}$. The theoretical risk of $\omega$ is then $R(\omega)=\mathbb{E}_{\mathbf{x}\sim\mathcal{D}}[\ell(\omega,\mathbf{x})]$, and its empirical counterpart is $\widehat{R}_{\mathcal{S}}(\omega) = \frac{1}{n} \sum_{i=1}^n \ell(\omega,\mathbf{x}_i)$. We let $\mathcal{M}_1(\Omega)$ denote the set of probability measures defined on $\Omega$ given some suitable $\sigma$-algebra. The primary focus of PAC-Bayes is to study the generalisation ability of random hypotheses $Q$ as measured by the gap between their average test risk $\mathbb{E}_{\omega\sim Q}[R(\omega)]$ and their average training risk $\mathbb{E}_{\omega\sim Q}[\widehat{R}_{\mathcal{S}}(\omega)]$. The following bound, due to McAllester \citep[the formulation is taken from][]{guedj2019primer}, provides a control on the generalisation gap valid for any `posterior' $Q$ using a data-independent prior distribution $\Pi$:
\begin{thm}
\label{thm-McAll-intro}
Let $\delta\in(0,1)$ and $\Pi\in\mathcal{M}_1(\Omega)$. Then with probability at least $1-\delta$ over $\{\mathbf{x}_1$,...,$\mathbf{x}_n\}\sim\mathcal{D}^n$, we have for any distribution $Q\in\mathcal{M}_1(\Omega)$:
$$ 
\mathbb{E}_{\omega\sim Q}[R(\omega)] \leq \mathbb{E}_{\omega\sim Q}[\widehat{R}_{\mathcal{S}}(\omega)] + \sqrt{\frac{\mathrm{KL}(Q\|\Pi)+\log\left(\frac{2\sqrt{n}}{\delta}\right)}{2n} } .
$$
\end{thm}
This PAC-Bayes bound is empirical, in the sense that the generalisation bound only depends on known computable quantities that are data-dependent. Hence as long as the upper bound is tight, minimising the right-hand side mitigates overfitting and guarantees the learner to get a method that uses the training data only to simultaneously learn the posterior $Q$ and get a risk certificate valid on unseen data. 

Note that the prior $\Pi$ involved in the KL term can be any data-independent distribution in $\mathcal{M}_1(\Omega)$. Hence, it appears that the choice of the distribution is crucial in order to obtain a bound that is not vacuous and thus that does not ignore important properties of the data-generating distribution. To tighten the bound, it has been advised to split the dataset, to learn $\Pi$ on the first subset and to compute the bound on the other part, thus fulfilling the condition of the theorem requiring the prior to be data-independent \citep{Ambroladze2007,Germain2009,ParradoHernandezPACData2012,KDRoyNonVacuous2017,mhammedi19pac,TCertificatesUCL2021,KDDataPAC2021,perezortiz2021learning}. The posterior can still be learnt on the entire data set.  

This prior learning step is even crucial for obtaining non-vacuous bounds on complex models such as deep neural networks. Indeed, many unsuccessful attempts have been made over the years in order to quantify the ability of overparameterised neural networks to generalise though achieving zero training error when trained with SGD \citep{ZhangDL2017}, and the first non-vacuous generalisation bounds in the modern deep learning regime were obtained in \citep{KDRoyNonVacuous2017} by deriving a PAC-Bayes bound starting from the solution produced by SGD, which is also chosen as the learnt prior mean. It turns out that although the weights change throughout the posterior learning stage, the SGD solution remains very close to the mean of the stochastic neural network, justifying then the use of PAC-Bayes bounds as a mean to derive guarantees on the SGD solution itself. Since then, several works used the same machinery to compute non-vacuous generalisation bounds for stochastic deep neural networks \citep{KDRoyDP2018,letarte19dichotomize,TCertificatesUCL2021,KDDataPAC2021,biggs2020differentiable,clerico2022}.

Obtaining such bounds for deterministic neural networks is far from being straightforward, and often requires a costly derandomising\footnote{Derandomising strategies have long been of interest in PAC-Bayes. Bounds holding with high probability
over a sampled predictor (directly drawn from the PAC-Bayes posterior) appear \emph{e.g.} in \citet{CatoniThermo}, \cite{AB2013,guedj2013}.} step. It can be done quite easily in very specific settings, as for instance in \cite{Germain2009} where the authors exploit the linearity of the hypotheses, but is very difficult to extend to the general case, see \emph{e.g.} \cite{Neyshabur2018,NagarajanKolter2019deterministic,NagarajanKolter2019uniform,BiggsBen2021}. Note that some generic derandomising techniques have been investigated recently, providing guarantees over one single hypothesis instead of the classical averaged analysis, and involving either the pointwise ratio $\log(Q(\omega)/\Pi(\omega))$ \citep{OmarDerandom2020} or an alternative divergence such as the Renyi \citep{Viallard2021} instead of the KL divergence.

We end this section by mentioning another recent PAC-Bayes bound referred to as the quadratic bound \citep{TCertificatesUCL2021}:
\begin{thm}
\label{thm-quad-intro}
Let $\delta\in(0,1)$ and $\Pi\in\mathcal{M}_1(\Omega)$. Then with probability at least $1-\delta$ over $\{\mathbf{x}_1$,...,$\mathbf{x}_n\}\sim\mathcal{D}^n$, we have for any distribution $Q\in\mathcal{M}_1(\Omega)$:
\begin{align*}
\mathbb{E}_{\omega\sim Q}&[R(\omega)] \leq \Bigg( \sqrt{\frac{\mathrm{KL}(Q\|\Pi)+\log(\frac{2\sqrt{n}}{\delta})}{2n}} \\ & + \sqrt{ \mathbb{E}_{\omega\sim Q}[\widehat{R}_{\mathcal{S}}(\omega)] + \frac{\mathrm{KL}(Q\|\Pi)+\log(\frac{2\sqrt{n}}{\delta})}{2n} } \Bigg)^2 .
\end{align*} 
\end{thm}
Similarly to McAllester's bound, this PAC-Bayes quadratic bound holds uniformly over all $Q$ and may be optimised with respect to $Q$. This bound is significantly tighter than McAllester's bound when the test loss is very small, and has been shown to provide tight risk certificates in practice when computed on real-life datasets \citep{TCertificatesUCL2021,perezortiz2021learning}.

\section{RECONSTRUCTION GUARANTEES FOR VAES}\label{sec:theory}

We adopt in this section a PAC-Bayesian approach on the VAE structure, both for computing generalisation bounds and for learning the related learning objective. The term pseudo-VAE refers to the structure learnt by any objective, whether that of the exact VAE, that of a $\beta$-VAE or that of a PAC-Bayes objective which we present in this section.

We consider a dataset $\mathcal{S}=\{\mathbf{x}_1$,\dots,$\mathbf{x}_n\}$ composed of binary data, typically images, from an unknown distribution $\mathcal{D}$. We use a Gaussian encoder and a standard Bernoulli conditional likelihood in the decoder as detailed in \eqref{variational} and \eqref{model}. Here, $\omega=(\phi,\theta)$, and the reconstruction loss $\ell(\phi,\theta,\mathbf{x})$ is obtained via rescaling $- \mathbb{E}_{q_{\phi}(\mathbf{z}|\mathbf{x})} \left[\log p_{\theta}(\mathbf{x}|\mathbf{z}) \right]$ where $p_{\theta}(\mathbf{x}|\mathbf{z})$ is a truncated version of the conditional likelihood, so that the loss is bounded with range $[0,1]$. Then, the theoretical $R(\phi,\theta)=\mathbb{E}_{\mathbf{x}\sim\mathcal{D}}[\ell(\phi,\theta,\mathbf{x})]$ and empirical $\widehat{R}_{\mathcal{S}}(\phi,\theta) = \frac{1}{n} \sum_{i=1}^n \ell(\phi,\theta,\mathbf{x}_i)$ reconstruction losses measure the quality of the reconstruction of any variant of the VAE whose encoder and decoder weights are respectively $\phi$ and $\theta$.

We now present our main results, which are generalisation bounds on the reconstruction gap. The first one, dedicated to designing the learning objective, is a bound on average over the VAE parameters, while the second one is a derandomised PAC-Bayes bound used to evaluate the upper bound. The first bound is given in the following theorem.
\begin{thm}
\label{thm-obj}
Let $\delta\in(0,1)$, $(\phi^0,\theta^0)$ and $\sigma^2_\theta>0$, $\sigma^2_\phi>0$. With probability at least $1-\delta$ over $\{\mathbf{x}_1$,...,$\mathbf{x}_n\}\sim\mathcal{D}^n$, we have for any $(\phi,\theta)$, for any $(s_\phi^2,s_\theta^2)$:
\begin{align*}
&\mathbb{E}_{\mathcal{N}(\phi,s^2_\phi I),\mathcal{N}(\theta,s^2_\theta I)}[R(\tilde\phi,\tilde\theta)] \leq \mathbb{E}_{\mathcal{N}(\phi,s^2_\phi I),\mathcal{N}(\theta,s^2_\theta I)}[\widehat{R}_{\mathcal{S}}(\tilde\phi,\tilde\theta)] \\
& + \sqrt{\frac{\|\phi-\phi^0\|_2^2}{4\sigma^2_\phi n} +\frac{N_\phi\left(\frac{s^2_\phi}{\sigma^2_\phi} + \log(\frac{\sigma_\phi^2}{s_\phi^2}) - 1\right)}{4n} + \frac{\log(\frac{2\sqrt{n}}{\delta})}{2n}} \\
& + \sqrt{\frac{\|\theta-\theta^0\|_2^2}{4\sigma^2_\theta n} +\frac{N_\theta\left(\frac{s^2_\theta}{\sigma^2_\theta} + \log(\frac{\sigma_\theta^2}{s_\theta^2}) - 1\right)}{4n} + \frac{\log(\frac{2\sqrt{n}}{\delta})}{2n}} ,
\end{align*}
where $N_\phi$ and $N_\theta$ are respectively the encoder and decoder neural networks size.
\end{thm}
The risks $R(\tilde\phi,\tilde\theta)$ and $\widehat{R}_{\mathcal{S}}(\tilde\phi,\tilde\theta)$ in the inequality are averaged over the random parameters $\tilde\phi$ and $\tilde\theta$ following Gaussians centered at the VAE parameters $\phi$, $\theta$ with respective variances $s_\phi^2$, $s_\theta^2$. The bound serves as a learning objective for both the VAE parameters $\phi$, $\theta$ and the corresponding variance levels $s_\phi^2$, $s_\theta^2$, and contains separate terms involving each of them.

The first quantity that controls the generalisation gap of the PAC-Bayes bound in \Cref{thm-obj} is the squared Euclidean distance between the learnt encoder weight $\phi$ (respectively decoder weight $\theta$) and a given prior mean encoder weight  $\phi^0$ (respectively prior mean decoder weight $\theta^0$), which can be chosen for instance as the starting point of the algorithm. This gives a PAC-Bayes objective involving this Euclidean distance as a measure of generalisation where the prior means can be learnt ex ante, which amounts here to learning $(\phi^0,\theta^0)$ using a subset of the data before training the VAE parameters $(\phi,\theta)$. The algorithm output $(\phi,\theta)$ will then stay close to the initialisation $(\phi^0,\theta^0)$, thus achieving a tight randomised bound and leading to good generalisation.

While the Euclidean metric in the bound involving the VAE parameters gives a quantitative way to predict whether a given pseudo-VAE can perform well on test data---if the bound is tight, then the VAE will generalise well to test data on average---the randomness also provides further insights on the regularisation effect of VAE objectives and may be used to answer the question: how could such PAC-Bayes objectives be more likely to generalise than standard VAEs? Indeed, although large values of the noise variances may seem irrelevant as leading to bounds on average that no longer apply to the VAE parameters directly, focusing on the noise level may reflect some notion of flatness \citep{HochreiterSchmidhuber1997}, which is known to be closely correlated to generalisation \citep{Keskar2017Flat,Rangamani2019} even though the exact relationship is still debated \citep{DinhSharp2017}. Some recent works have explored the links between PAC-Bayes and flatness \citep{KDRoyNonVacuous2017,Tsuzuku2020flat,Pitas2020Dissecting}. Typically, the more noise we can add without hurting the training reconstruction loss at a given minimum, the flatter is this achieved minimum. Hence, injecting a small amount of noise should force the minimisation procedure to end in flat regions of the reconstruction landscape, and learning the noise level during the training stage could balance flatness and parameters scales while preserving the algorithm from computational difficulties. Formalising this idea leads us to consider our randomised PAC-Bayes bound on the test reconstruction error to be minimised.

We present now a derandomised bound, which is a key point in order to evaluate the performance of the learnt strategy itself, in contrast to the previous standard averaged PAC-Bayes bound.
\begin{thm}
\label{thm-main}
Let $\delta\in(0,1)$, $(\phi^0,\theta^0)$ and $\sigma^2_\theta>0$, $\sigma^2_\phi>0$. Then with probability at least $1-\delta$ over both $\mathcal{S}=\{\mathbf{x}_1$,...,$\mathbf{x}_n\}\sim\mathcal{D}^n$, and $\varepsilon_\phi\sim\mathcal{N}(0,\sigma^2_\phi I)$, $\varepsilon_\theta\sim\mathcal{N}(0,\sigma^2_\theta I)$:
\begin{multline}
\label{bound-noisy}
\textrm{kl}\left(R(\phi+\varepsilon_\phi,\theta+\varepsilon_\theta)\|\widehat{R}_{\mathcal{S}}(\phi+\varepsilon_\phi,\theta+\varepsilon_\theta)\right) \\
\leq \frac{\|\phi-\phi^0+\varepsilon_\phi\|_2^2-\|\varepsilon_\phi\|_2^2}{2\sigma^2_\phi n} + \frac{\|\theta-\theta^0+\varepsilon_\theta\|_2^2-\|\varepsilon_\theta\|_2^2}{2\sigma^2_\theta n}  \\ + \frac{\log(2\sqrt{n}/\delta)}{n}  ,
\end{multline}
where $(\phi,\theta)$ is the output of the algorithm minimising the bound in \ref{thm-obj} given the dataset $\mathcal{S}$, and where $\textrm{kl}$ is the binary Kullback-Leibler divergence
$$
\textrm{kl}(q\|p) = q\log\left(\frac{q}{p}\right) + (1-q)\log\left(\frac{1-q}{1-p}\right)
$$
for any $q,p\in(0,1)$.
\end{thm}

The proof is deferred to the supplementary material.

Note that it is also possible to obtain a bound on the $\textrm{kl}$ generalisation gap that does not depend on the Gaussian noises $\varepsilon_\phi$, $\varepsilon_\theta$ at the price of a slightly larger constant in front of the squared Euclidean distances. Indeed, using proof techniques from \cite{Viallard2021}, the bound in Theorem \ref{thm-main} is replaced by:
\begin{align*}
\frac{\|\phi-\phi^0\|_2^2}{2\sigma^2_\phi n} + \frac{\|\theta-\theta^0\|_2^2}{2\sigma^2_\theta n} + \frac{\log(2\sqrt{n}/\delta)}{2n} .
\end{align*}

The high probability bound over the noise means that inequality \eqref{bound-noisy} holds for any encoder/decoder weights that is perturbed with a Gaussian distribution. This comes from the derandomisation step used within the PAC-Bayes machinery, and makes the bound stand for a unique (but perturbed) pair of encoder/decoder weights rather than for a stochastic one as it is usually the case in PAC-Bayes. The risk is then individual and not averaged over the Gaussian noise. A subsequent advantage is that the bound can be evaluated by drawing one sample only, whereas the randomised error needs to be estimated by Monte Carlo, which adds another term in the bound depending on the number of samples drawn for approximating the randomised error. In practice, when the noise standard deviations $\sigma_\phi$ and $\sigma_\theta$ are small, \emph{e.g.} $10^{-2}$, the standard deviations of the reconstruction losses over random Gaussian perturbations are in order $10^{-2}$ as well, and the bound can simply be approximated as follows: with high probability over the drawing of the dataset and the Gaussian perturbation, we have for any $(\phi,\theta)$,
\begin{multline*}   \textrm{kl}\left(R(\phi,\theta)\|\widehat{R}_{\mathcal{S}}(\phi,\theta)\right) \leq \\
\frac{\|\phi-\phi^0+\varepsilon_\phi\|_2^2-\|\varepsilon_\phi\|_2^2}{2\sigma^2_\phi n} + \frac{\|\theta-\theta^0+\varepsilon_\theta\|_2^2-\|\varepsilon_\theta\|_2^2}{2\sigma^2_\theta n} \\ + \frac{\log(2\sqrt{n}/\delta)}{n}  .
\end{multline*}
We can easily obtain a tight bound on the test reconstruction error directly from \eqref{bound-noisy} by using a numerical inversion procedure on the binary $\textrm{kl}$. The procedure is inspired by \cite{KDRoyNonVacuous2017,TCertificatesUCL2021} and adapted to our setting in the supplementary material.

To summarise, we considered two different bounds with two different perspectives. \Cref{thm-obj} is a randomised inequality designed to learn the noise level so that it encourages the learning process to end in flat regions with small generalisation error, while \Cref{thm-main} is a high probability bound over the noise which is used to evaluate the final bound for small fixed values of the noise so that the bound can be considered as deterministic. Note that learning is no longer self-certified, but the principles driving this approach could still lead to good guarantees.

Our PAC-Bayes approach can be summarised as: 
\begin{enumerate}
\item[(i)] First learn $(\phi^0,\theta^0)$ (more details on the procedure in \Cref{sec:experiments});
\item[(ii)] Then learn $(\phi,\theta)$ along with $(s^2_\phi,s^2_\theta)$. This can be done by minimising the bound in \Cref{thm-obj} using SGD on the entire dataset;
\item[(iii)] Finally evaluate the bound at $(\phi,\theta)$. This can be done by inverting the $\textrm{kl}$ bound in \Cref{thm-main}.
\end{enumerate}

\section{EXPERIMENTS}\label{sec:experiments}

We designed some experiments on MNIST and Omniglot in order to:
\begin{itemize}
   \item evaluate the generalisation ability of both the $\beta$-VAE and PAC-Bayes objectives in terms of reconstruction,
   \item compute generalisation bounds for the aforementioned learning strategies. 
\end{itemize}

\subsection{Experimental setup}

The architecture of both the encoder and decoder networks used for all experiments are feedforward neural networks with 3 layers each (excluding the ‘input layer’), 400 units per hidden layer, and ReLU activations. The dimensionality of the latent space is $50$. We trained our
models on MNIST using the standard split composed of 60,000 training and 10,000 test datapoints, while on Omniglot we use the split in \cite{burda2015importance} with 24,345 training and 8,070 test datapoints. We use the Adam optimiser \citep{kingma2014adam} with learning rate 1e-3 and minibatch size 100. For the PAC-Bayes objectives, in order to obtain a bounded loss function, we clamp the output layer $\omega_\theta(\mathbf{z})$ of the decoder between $p_{\text{min}}$ and $1-p_{\text{min}}$ similar to \cite{dziugaite2018datadependent, TCertificatesUCL2021}. We find in practice that a value of $p_{\text{min}}=\text{5e-3}$ works well in practice and values higher than $10^{-2}$ can cause degradation to the reconstruction quality. For all procedures, we train the VAE for approximately 500k iterations including prior learning (see below).

\textbf{Prior learning.} The key ingredient in PAC-Bayes in order to obtain learning procedures with tight bounds that generalise well lies in learning the prior. The standard method to learn the prior mean $(\phi^0,\theta^0)$ is empirical risk minimisation via SGD. \cite{TCertificatesUCL2021} recommended the use of dropout without which the learnt prior is prone to overfitting. Such a procedure is not adapted to our structure, as minimising the training reconstruction loss in isolation boils down to learning a deterministic autoencoder, which should be avoided. In order to address this issue, we learn $(\phi^0,\theta^0)$ as a $\beta$-VAE objective minimiser for $\beta>0$, using SGD with and without dropout. In turns out that dropout is particularly important for smaller values of $\beta$. We tested in our experiments both data dependent priors using a $\beta$-VAE, and data independent priors centred at zero or a random parameter value drawn from a clamped normal distribution (with standard deviation $1/\sqrt{n_{\text{input features}}}$).

We also performed a grid sweep over prior standard deviations $(\sigma_\phi,\sigma_\theta)$ in $[0.005,0.01,0.03,0.05]$. We observed that the best value of standard deviations is usually $0.01$, which seems to provide a good tradeoff between stability and exploration.

\textbf{KL attenuating trick.} A common issue encountered in PAC-Bayes is that the training loss is often dominated by the KL term, which tends to make the optimisation program focus on minimising the KL \citep{Lever2013,Blundell2015,KDRoyDP2018,TCertificatesUCL2021} and makes the posterior not move far from the prior. A way to address this problem is to use the so-called KL attenuating trick that consists in downweighting the KL in the objective \citep{Blundell2015,TCertificatesUCL2021}. Here, it means that we multiply the Euclidean distances by a small factor during training. This factor is set to 0.0001 in the experiments when using the KL attenuating trick. 

\textbf{Reparameterisation.} We used the reparameterisation $\rho_\phi=\log(s_\phi)$, $\rho_\theta=\log(s_\theta)$ and optimise $(\rho_\phi,\rho_\theta)$ instead of $(s_\phi,s_\theta)$ so that the standard deviations always stay non-negative.

\subsection{Results}

\begin{figure*}[t]
\begin{centering}
\vspace{-2.5em} 
\subfloat[]{\includegraphics[width=8cm]{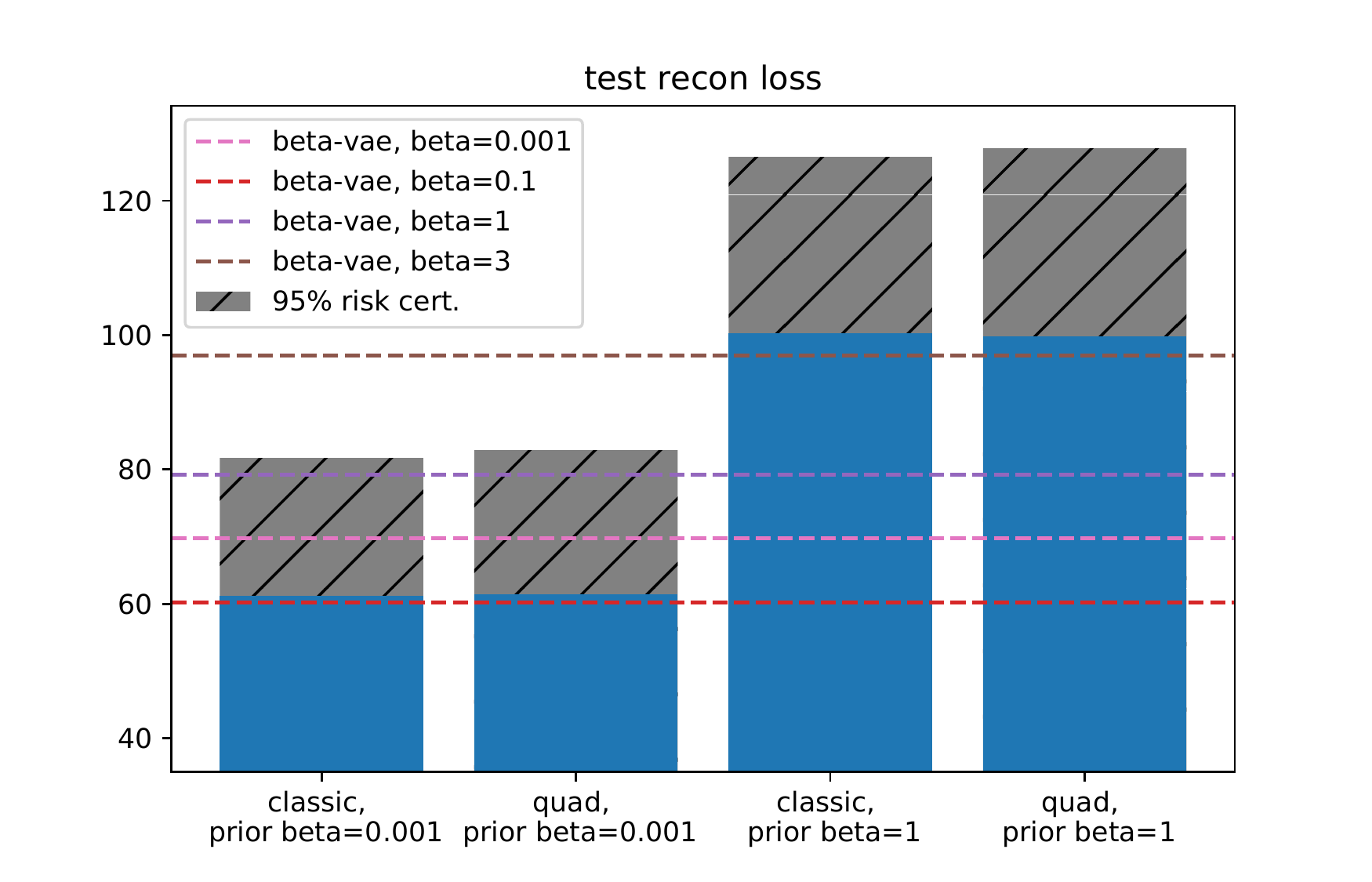}}\subfloat[]{\includegraphics[width=8cm]{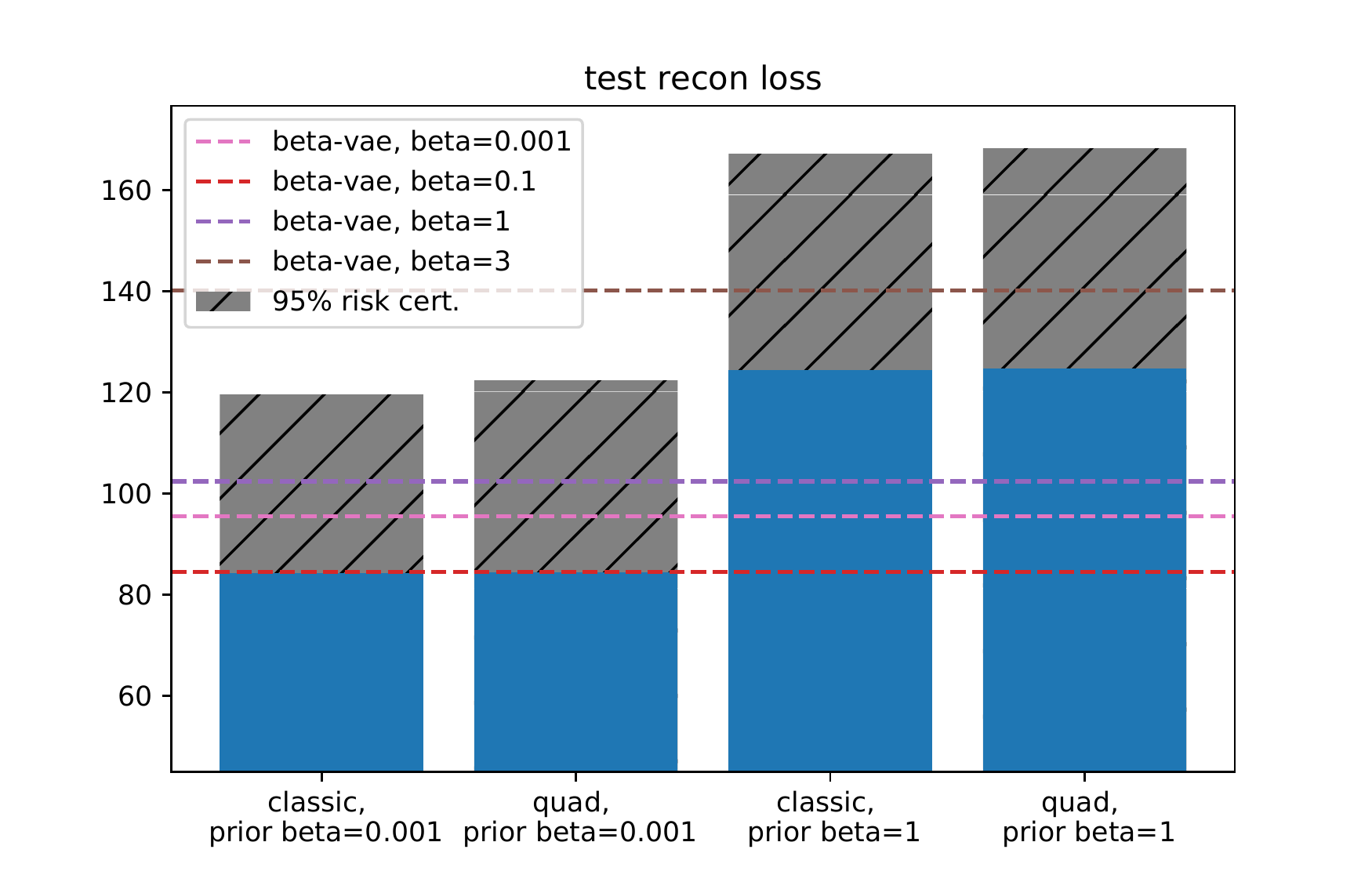}}
\par\end{centering}
\caption{Test reconstruction loss and PAC-Bayes bounds on the (a) binarised MNIST; (b) Omniglot datasets. Models trained using PAC-Bayes objectives can achieve the same level of test reconstruction error as the best $\beta$-VAE model with $\beta=0.1$, while also providing tight risk certificate. }
\end{figure*}

\begin{figure*}[t]
\begin{centering}
\vspace{-1.25em} 
\subfloat[]{\includegraphics[height=5cm]{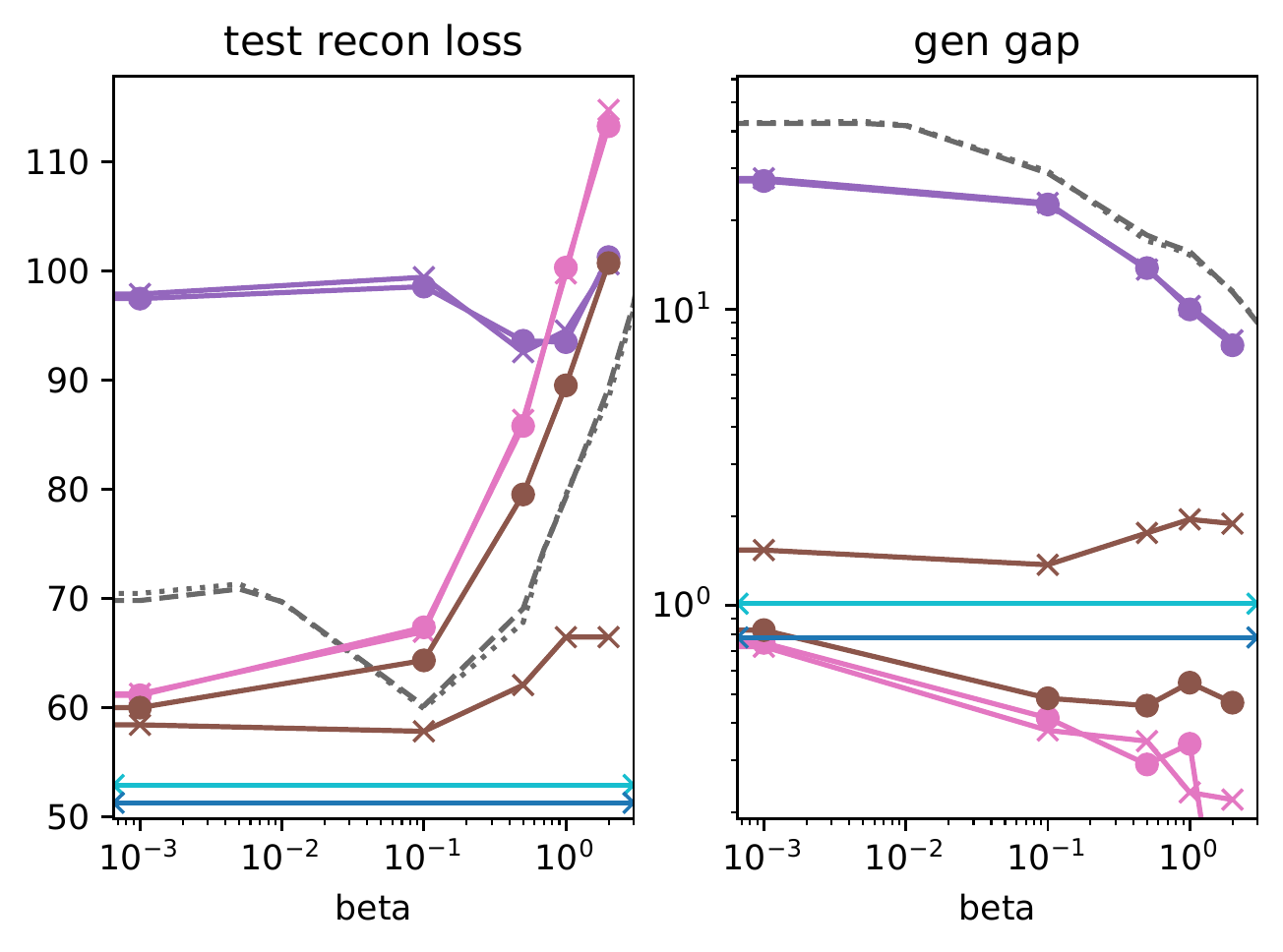}}\subfloat[]{\includegraphics[height=5cm]{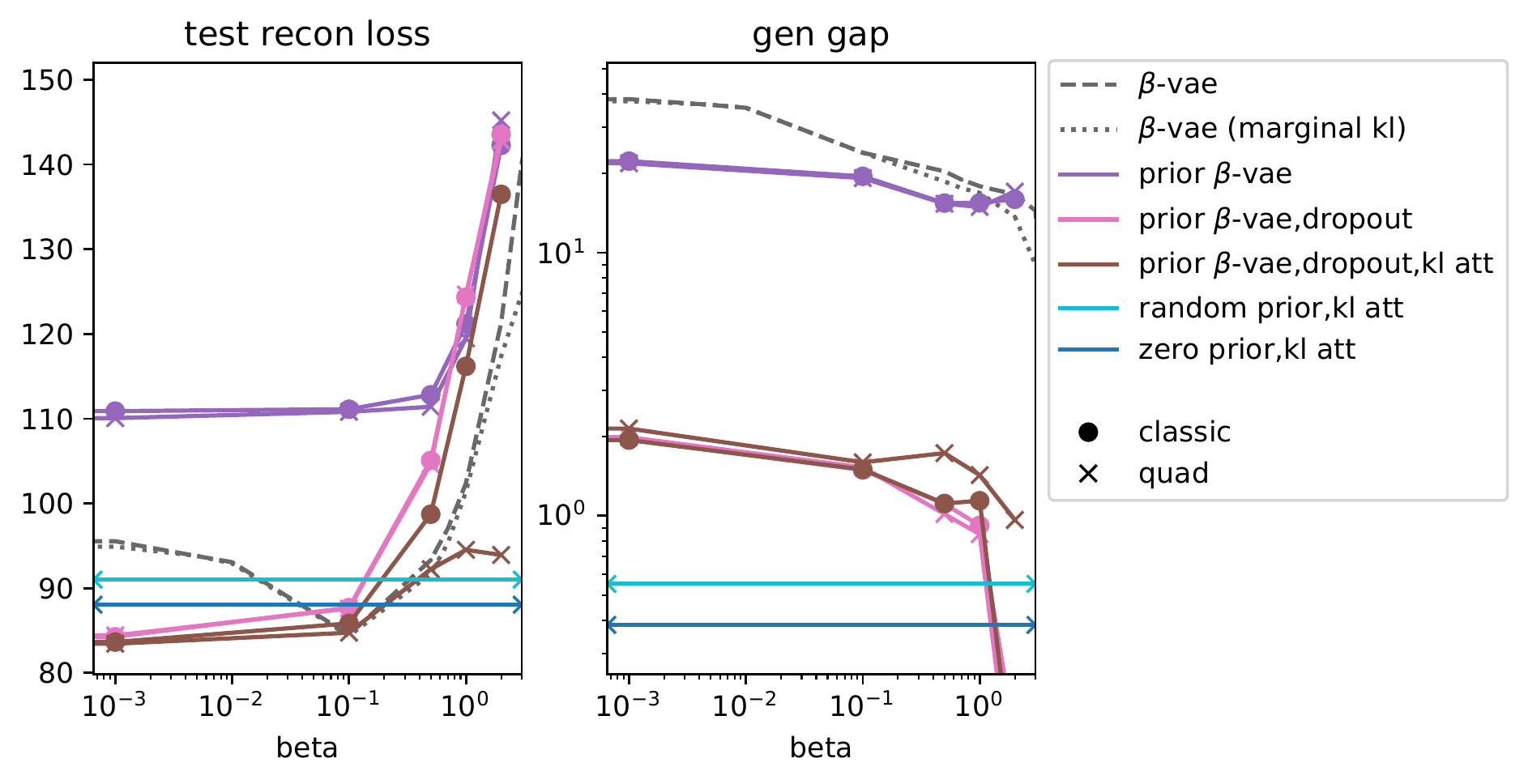}}
\par\end{centering}
\caption{Test reconstruction vs generalisation on the (a) binarised MNIST; (b) Omniglot datasets. All models trained using PAC-Bayes objectives (solid lines) achieve significantly better generalisation ability than $\beta$-VAEs (dashed lines), including the marginal KL $\beta$-VAE baseline \citep{ELBOSurgery2016} which uses marginal KL as the regulariser. Models with best settings also outperform $\beta$-VAEs consistently in terms of test reconstruction. }

\end{figure*}

The main findings from our experiments can be summarised as follows: i) the PAC-Bayes bounds are vacuous for $\beta$-VAE objectives but can be made non-vacuous when using the PAC-Bayes objectives; ii) the $\beta$-VAE objectives are much more prone to overfitting than PAC-Bayes objectives, while achieving at most comparable test reconstruction performance. This suggests, in line with \cite{Bozkurt2021} findings, that the KL term is not a particularly relevant measure of generalisation in terms of reconstruction and that the good empirical performance of standard VAEs can largely be attributed to their known memorisation property.

Figure 1 displays generalisation bounds from different PAC-Bayes learning objectives without KL attenuation, each objective corresponding to a value of $\beta$ when learning the prior mean using a $\beta$-VAE with dropout. We rescale all the reconstruction losses and certificates so that they can be compared to the loss values reported in the literature. The blue area corresponds to the test reconstruction loss of the learnt procedure, while the shaded area on top corresponds to the bound. The smallest the shaded area, the tightest the bound. We can see that the computed bounds are reasonably tight. The horizontal dashed lines represent the test reconstruction loss achieved by several $\beta$-VAEs.

Figure 2 shows both the test reconstruction loss and the generalisation gap for a wide range of $\beta$ values and training objectives: “prior $\beta$-VAE” corresponds to PAC-Bayes objectives with starting point $(\phi_0,\theta_0)$ learnt using a $\beta$-VAE, while the starting points of “random prior” and “zero prior” curves are respectively drawn randomly using a Gaussian or simply put to $(\phi_0,\theta_0)$. “dropout” means that dropout is used when learning $(\phi_0,\theta_0)$ using a $\beta$-VAE, while “kl att” means that the kl-attenuating trick has been used when learning $(\phi,\theta)$ using the bound in \Cref{thm-obj}. We insist on the fact that we use two different notions of prior here: the prior $p(\mathbf{z})$ used in standard VAEs which is always selected as a standard multivariate Gaussian,
and the priors placed over the parameters $\phi$ and $\theta$ which are chosen to be Gaussian centred at $(\phi_0,\theta_0)$ respectively.

Note that even though the generalisation ability is the main focus of this paper, this notion is not interesting per se. Indeed, a procedure that would learn to reconstruct images uniformly at random would generalise very well since the training and test losses would be exactly the same. Hence, we are looking for strategies with competitive test reconstruction loss while achieving a small generalisation gap, and Figure 2 is particularly enlightening from this perspective. We observe that when using the KL attenuating trick, the quadratic bound is better for test reconstruction than McAllester's bound when the prior mean is learnt using a $\beta$-VAE - whatever the value of $\beta$ - but generalises worse, while without KL attenuation they are comparable for both reconstructing test data and generalising.

Interestingly, learning the prior mean does not appear to be crucial here, in sharp contrast to the general PAC-Bayes workflow. For instance, when we set $\phi^0$ and $\theta^0$ to 0 and then train the PAC-Bayes objective using dropout and the KL attenuating trick, we observe that the obtained test reconstruction and generalisation gap are both extremely competitive on both datasets. When learning a prior using a $\beta$-VAE, this behavior seems nonetheless to deteriorate quickly as the $\beta$ value increases, with a very bad reconstruction and an almost vanishing generalisation gap, which seems to suggest that there is underfitting.

Regarding computational considerations, the KL attenuating trick seems to improve significantly the test reconstruction loss as observed on Figure 2. Unfortunately, the effect of downweighting the regularisation term is that learnt parameters $(\phi,\theta)$ are allowed to move far from the prior mean parameters $(\phi^0,\theta^0)$, which results in vacuous certificates. Nevertheless, we remark that those vacuous certificates do not correlate with bad generalisation, as the generalisation gap is still small for all datasets, which might invite to consider a possible tradeoff between test reconstruction performance and tightness of the bounds without depreciating the generalisation ability. This behavior is common in PAC-Bayes, and particularly strong here. Indeed, we observe that the PAC-Bayes VAE with the KL attenuating trick and dropout never significantly worsen the performance of the prior mean it is based upon. For instance, when $(\phi^0,\theta^0)$ is learnt using a $\beta$-VAE, then the performance of $(\phi,\theta)$ learnt using the quadratic bound is almost always better than the performance of $(\phi^0,\theta^0)$. However, this behavior does not occur without the KL attenuating trick, which tends to say that larger test reconstruction loss is often a price to pay to get tight risk certificates, and conversely.

\section{CONCLUSION}

In this paper, we derived generalisation bounds for VAEs and showed that VAE objectives can be designed in such a way as to achieve a competitive reconstruction loss while generalising well. This discussion on a PAC-Bayes objective not only puts into perspective the choice of the KL term as a regulariser, but also calls for principled regularising objectives (with guarantees) that would prevent overfitting in general.

An interesting further remark is that the prior $p(\mathbf{z})$ is not involved in our PAC-Bayes objective. Hence, this suggests that learning the prior $p(\mathbf{z})$ \emph{after} learning the encoder and the decoder, as recommended in several works \citep{ELBOSurgery2016,VampPrior2018} could help achieving significantly better sampling properties without depreciating the generalisation ability of the learnt architecture. It is particularly interesting as preliminary experiments seem to show that a PAC-Bayes VAE whose prior mean $(\phi^0,\theta^0)$ is learnt with a $\beta$-VAE, with dropout and without KL attenuation, inherits part of the sampling ability of the $\beta$-VAE it is based on. We see this as a promising avenue for future research.

Another question of interest is the choice of the learning procedure for $(\phi^0,\theta^0)$. Indeed, basic strategies that do not even learn these parameters can reach highly competitive performance, which is in sharp contrast to the classical predictive machine learning setting. The benefit of learning $(\phi^0,\theta^0)$ could then lie in its sampling properties rather than its reconstruction properties. We leave this investigation to future works.

\paragraph{Societal impact.} Due to the theoretical nature of our work, we do not foresee immediate societal consequences, although we hope through contributing a better theoretical understanding, the present paper can foster a more informed use of VAEs by practitioners.

\subsubsection*{Acknowledgements}

Badr-Eddine Chérief-Abdellatif, Benjamin Guedj and Arnaud Doucet acknowledge support of the UK Defence Science and Technology Laboratory (DSTL) and EPSRC under grants EP/R018693/1 and EP/R013616/1, as part of the collaboration between US DOD, UK MOD and UK EPSRC under the Multidisciplinary University Research Initiative.
Benjamin Guedj acknowledges partial support from the French National Agency for Research, grants ANR-18-CE40-0016-01 and ANR-18-CE23-0015-02. Arnaud Doucet acknowledges partial support from the EPSRC CoSInES grant EP/R034710/1.

\bibliographystyle{plainnat}
\bibliography{biblio}


\clearpage
\appendix

\thispagestyle{empty}

\onecolumn \makesupplementtitle


\section{PROOF OF THEOREM \ref{thm-main}}

We first recall \Cref{thm-main}. We denote $(\phi,\theta)$ the parameter learnt from the training dataset $\mathcal{S}$ using the procedure proposed in Section 3 of the paper. 

\begin{thm}
\label{thm-main2}
Let $\delta\in(0,1)$, $(\phi^0,\theta^0)$, and $\sigma^2_\theta>0$, $\sigma^2_\phi>0$. 
Then we have with probability at least $1-\delta$ over both $\mathcal{S}=\{\mathbf{x}_1,...,\mathbf{x}_n\}\sim\mathcal{D}^n$, and $\varepsilon_\phi\sim\mathcal{N}(0,\sigma^2_\phi I)$, $\varepsilon_\theta\sim\mathcal{N}(0,\sigma^2_\theta I)$:
\begin{multline*}
\textup{kl}\left(\widehat{R}_{\mathcal{S}}(\phi+\varepsilon_\phi,\theta+\varepsilon_\theta)\|R(\phi+\varepsilon_\phi,\theta+\varepsilon_\theta)\right) \leq \frac{\|\phi-\phi^0+\varepsilon_\phi\|_2^2-\|\varepsilon_\phi\|_2^2}{2\sigma^2_\phi n} + \frac{\|\theta-\theta^0+\varepsilon_\theta\|_2^2-\|\varepsilon_\theta\|_2^2}{2\sigma^2_\theta n} + \frac{\log(2\sqrt{n}/\delta)}{n} ,
\end{multline*}
where $(\phi,\theta)$ is the output of the algorithm given the dataset $\mathcal{S}$.
\end{thm}

\begin{proof}
The proof exploits a result from \cite{OmarDerandom2020}. We use the following notations: $\mathcal{S}=\{\mathbf{x}_1,...,\mathbf{x}_n\}$ for the training dataset, $h=(\tilde\phi,\tilde\theta)$ for the parameter, $Q^0(\cdot)=\mathcal{N}(\phi^0,\sigma^2_\phi I)\otimes \mathcal{N}(\theta^0,\sigma^2_\theta I)$ for the data-free prior, $Q_{\mathcal{S}}(\cdot)=\mathcal{N}(\phi,\sigma^2_\phi I)\otimes \mathcal{N}(\theta,\sigma^2_\theta I)$ for the posterior (given the dataset $\mathcal{S}$ and the pair $(\phi,\theta)$ learnt from $\mathcal{S}$), and:
$$
f(\mathcal{S},h) = n \cdot \textrm{kl}\left(\hat{R}_{\mathcal{S}}(\tilde\phi,\tilde\theta) \| R(\tilde\phi,\tilde\theta) \right)  .
$$
Hence, according to Theorem 1(i) from \cite{OmarDerandom2020}, we have with probability at least $1-\delta$ over both $\mathcal{S}=\{\mathbf{x}_1$,...,$\mathbf{x}_n\}\sim\mathcal{D}^n$, and $\tilde\phi\sim\mathcal{N}(\phi,\sigma^2_\phi I)$, $\tilde\theta\sim\mathcal{N}(\theta,\sigma^2_\theta I)$:
$$   
\textrm{kl}\left(\hat{R}_{\mathcal{S}}(\tilde\phi,\tilde\theta) \| R(\tilde\phi,\tilde\theta) \right) \leq \frac{1}{n} \left\{ \log\left(\frac{Q_{\mathcal{S}}(\tilde\phi,\tilde\theta)}{Q^0(\tilde\phi,\tilde\theta)}\right) + \log\left(\frac{\zeta}{\delta}\right) \right\} ,
$$
where $\zeta=\mathbb{E}_{\bar\phi \sim \mathcal{N}(\phi^0,s^2_\phi I),\bar\theta \sim \mathcal{N}(\theta^0,s^2_\theta I)}\left[ \exp \left( n \cdot \textrm{kl}\left(\hat{R}_{\mathcal{S}}(\bar\phi,\bar\theta) \| R(\bar\phi,\bar\theta) \right) \right) \right]$.  

We compute the first term in the right-hand-side:
\begin{align*}
\log\left(\frac{Q_{\mathcal{S}}(\tilde\phi,\tilde\theta)}{Q^0(\tilde\phi,\tilde\theta)}\right) & = \log\left(\frac{(2\pi\sigma^2_\phi)^{-N_\phi/2}\exp\left(-\frac{\|\tilde\phi-\phi^2\|_2}{2\sigma^2_\phi}\right)}{(2\pi\sigma^2_\phi)^{-N_\phi/2}\exp\left(-\frac{\|\tilde\phi-\phi^0\|_2^2}{2\sigma^2_\phi}\right)}\right) + \log \left(\frac{(2\pi\sigma^2_\theta)^{-N_\theta/2}\exp\left(-\frac{\|\tilde\theta-\theta\|_2^2}{2\sigma^2_\theta}\right)}{(2\pi\sigma^2_\theta)^{-N_\theta/2}\exp\left(-\frac{\|\tilde\theta-\theta^0\|_2^2}{2\sigma^2_\theta}\right)}\right) \\ 
 & = \frac{\|\tilde\phi-\phi^0\|_2^2-\|\tilde\phi-\phi\|_2^2}{2\sigma^2_\phi} + \frac{\|\tilde\theta-\theta^0\|_2^2-\|\tilde\theta-\theta\|_2^2}{2\sigma^2_\theta} ,
\end{align*}
where $N_\phi$ and $N_\theta$ are the respective lengths of $\phi$ and $\theta$. Moreover, $\zeta$ is upper bounded by $2\sqrt{n}$ according to \cite{Maurer2004}, which gives the final result when writing $\tilde\phi=\phi+\varepsilon_\phi$, $\tilde\theta=\theta+\varepsilon_\theta$ with $\varepsilon_\phi\sim\mathcal{N}(0,\sigma^2_\phi I_N)$, $\varepsilon_\theta\sim\mathcal{N}(0,\sigma^2_\theta I_N)$.
\end{proof}
  
\Cref{thm-obj} is a simple application of McAllester's bound (\Cref{thm-McAll-intro}) applied to $Q(\cdot)=\mathcal{N}(\phi,s^2_\phi I)\otimes \mathcal{N}(\theta,s^2_\theta I)$ and $\Pi(\cdot)=\mathcal{N}(\phi^0,\sigma^2_\phi I)\otimes \mathcal{N}(\theta^0,\sigma^2_\theta I)$. 

\section{THE $kl$ INVERSION PROCEDURE}
  
\Cref{thm-main} only provides a bound on the binary $\textrm{kl}$ divergence between the empirical and theoretical losses $\hat{R}_{\mathcal{S}}(\phi+\varepsilon_\phi,\theta+\varepsilon_\theta)$ and $R(\phi+\varepsilon_\phi,\theta+\varepsilon_\theta)$. It is possible to obtain a bound on the theoretical loss directly \citep{KDRoyNonVacuous2017,TCertificatesUCL2021}. We introduce:
$$
\forall p\in[0,1], \quad \forall c\geq 0, \quad \textrm{kl}^*(p|c) = \sup\left\{ q\in[p,1], \textrm{kl}(p\|q) \leq c \right\} ,
$$
which is well-defined and satisfies $q\leq \textrm{kl}^*(p|c)$ when $\textrm{kl}(p\|q)\leq c$. This leads to:
\begin{multline*}
R(\phi+\varepsilon_\phi,\theta+\varepsilon_\theta) \leq \textrm{kl}^*\bigg(\widehat{R}_{\mathcal{S}}(\phi+\varepsilon_\phi,\theta+\varepsilon_\theta) \bigg| \frac{\|\phi-\phi^0+\varepsilon_\phi\|_2^2-\|\varepsilon_\phi\|_2^2}{2\sigma^2_\phi n} + \frac{\|\theta-\theta^0+\varepsilon_\theta\|_2^2-\|\varepsilon_\theta\|_2^2}{2\sigma^2_\theta n} + \frac{\log(2\sqrt{n}/\delta)}{n}\bigg)
\end{multline*}
with probability at least $1-\delta$. 

Hence, to compute the upper bound, it is sufficient to evaluate  $\textrm{kl}^*$, which is a difficult task but can be done using a numerical approximation. We used the same implementation as in \cite{TCertificatesUCL2021}.

\end{document}